\pdfoutput=1
\documentclass{article}

\PassOptionsToPackage{square}{natbib}
\usepackage[preprint]{neurips_2026}

\usepackage[utf8]{inputenc}
\usepackage[T1]{fontenc}
\usepackage{hyperref}
\usepackage[figure]{hypcap}
\usepackage{url}
\mathchardef\UrlBigBreakPenalty=9000
\usepackage{booktabs}
\usepackage{amsmath}
\usepackage{amssymb}
\usepackage{amsfonts}
\usepackage{nicefrac}
\usepackage{microtype}
\usepackage{xcolor}
\usepackage{graphicx}
\usepackage{tabularx}
\usepackage{array}
\usepackage{threeparttable}
\usepackage{enumitem}
\usepackage{xspace}
\definecolor{linkblue}{HTML}{1C5CAB}
\hypersetup{
  colorlinks=true, linkcolor=linkblue, citecolor=linkblue, urlcolor=linkblue,
  pdftitle={Cross-Benchmark Transfer from RL on Agentic Coding Tasks},
  pdfauthor={Sushant Mehta, Logan Ritchie, Edwin Chen},
  pdfsubject={Reinforcement learning for agentic coding},
  pdfkeywords={reinforcement learning, RLVR, coding agents, GSPO, SWE-Bench Pro, Terminal-Bench, SWE-Marathon}
}

\newcommand{\pp}{\,pp\xspace}
\newcommand{\ra}{$\rightarrow$}
\newcommand{\code}[1]{\texttt{#1}}
\newcolumntype{Y}{>{\raggedright\arraybackslash}X}

\title{Cross-Benchmark Transfer from RL\\on Agentic Coding Tasks}

\author{%
  Sushant Mehta \qquad Logan Ritchie \qquad Edwin Chen\\
  Surge AI\\
  \texttt{research@surgehq.ai}
}

\begin{document}

\maketitle

\begin{abstract}
Coding agents often fail in the last mile: they build most of a feature but drop a requirement, test only the cases their implementation already handles, break behavior that was supposed to stay intact, or validate against an unchecked assumption. We ask whether reinforcement learning (RL) on expert-built agentic coding tasks closes this gap, and whether what the agent learns transfers beyond the training distribution. We post-train Kimi~K2.7~Code, a 1T-parameter (32B active) open-weight mixture-of-experts model, with RL alone on 1,700 tasks: 1,000 repository tasks graded by hidden fail-to-pass tests and by pass-to-pass tests of existing behavior, and 700 terminal tasks graded by expert-written hidden verifiers. The reward is the fraction of target checks passed and drops to zero if any pass-to-pass test fails. One epoch of GSPO on a rank-32 LoRA adapter improves pass@1 on each of the six external benchmarks we evaluated, across three agent harnesses: SWE-Bench~Pro (60.1\ra64.8), DeepSWE (31.0\ra43.4), Terminal-Bench~2.1 (67.4\ra82.0), Terminal-Bench~3 (1.4\ra12.1), Terminal-Bench~4 (0.0\ra7.6), and SWE-Marathon (5.0\ra25.0). Pooled over the five independent task sets (Terminal-Bench~4 revises Terminal-Bench~3), the improvement is significant ($p<0.001$), and it remains significant on the three sets released after the training data was collected ($p=0.004$); the model also improves under both harnesses never used in training. Median trajectories on DeepSWE and Terminal-Bench~3 are 24--35\% shorter in agent steps. The base model's failed DeepSWE runs are mostly near-misses, and on the tasks the trained model newly solves, paired trajectories show it avoiding each of the four failure modes above.
\end{abstract}

\section{Introduction}
\label{sec:intro}

Reinforcement learning with verifiable rewards (RLVR) is now a standard stage in training coding agents~\citep{deepseekr1,swerl2025,deepsweagentica2025,kimik2}. The usual evidence that a run worked is a higher score on a benchmark whose format the training data shares. That leaves two questions open. What did the model learn: domain knowledge, or something more general about carrying a task to completion? And does the improvement survive a change of benchmark, agent harness, and horizon? An agent that has absorbed one benchmark's conventions is worth less than one that has learned to finish realistic work.

\begin{figure}[t]
  \centering
  \includegraphics[width=\linewidth]{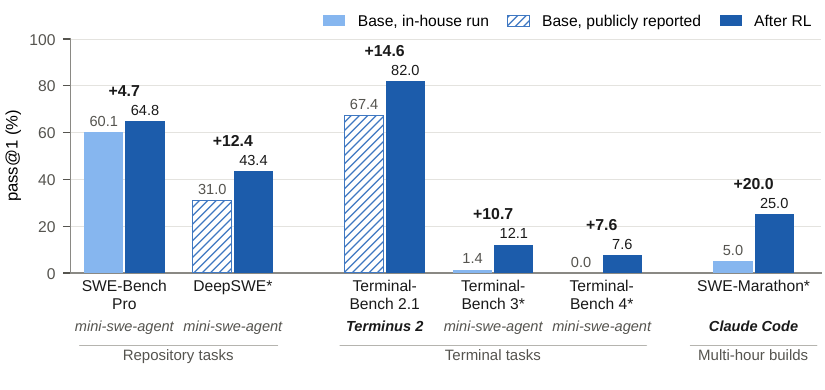}
  \caption{Pass@1 on six external benchmarks before and after RL on 1,700 expert-built coding tasks, grouped by task format, with SWE-Marathon's multi-hour builds shown separately. Bold numbers are gains in percentage points. Harnesses are in italics, in bold for the two not used in training. Hatched baselines are publicly reported scores for the same checkpoint and harness (DeepSWE: benchmark leaderboard; Terminal-Bench~2.1: Artificial Analysis); the others are in-house runs. Asterisks mark benchmarks released after the training data was collected; Terminal-Bench~4 revises Terminal-Bench~3.}
  \label{fig:results}
\end{figure}

We study both questions with one training run and six external benchmarks. We post-train Kimi~K2.7~Code~\citep{moonshot2026kimik27code} with RL alone on 1,700 expert-built agentic coding tasks from two families (\S\ref{sec:data}). Each of the 1,000 repository tasks pairs a real repository at a pinned commit with a change request; hidden fail-to-pass tests grade the change, and pass-to-pass tests guard existing behavior. Each of the 700 terminal tasks pairs a working environment with a deliverable graded by expert-written hidden verifiers. The recipe is deliberately simple, so that what the run tests is the tasks and their reward: Group Sequence Policy Optimization (GSPO)~\citep{gspo2025} on a rank-32 LoRA adapter~\citep{lora2022}, no supervised warm-up, no KL penalty, and a partial-credit reward, the fraction of satisfied checks, that drops to zero whenever a pass-to-pass test regresses (\S\ref{sec:method}).

We compare the base and trained models on SWE-Bench~Pro, DeepSWE, Terminal-Bench~2.1, 3, and~4, and SWE-Marathon, fixing one harness per benchmark, three harnesses in all (\S\ref{sec:eval}). The six cover both training formats, new task sources and domains, and a horizon absent from training: SWE-Marathon's multi-hour builds. Four were released after the training data was collected. The trained model improves on all six (\S\ref{sec:results}; Figure~\ref{fig:results}), by 4.7 to 20.0 percentage points (pp), including under both harnesses never used in training: Claude~Code on SWE-Marathon (1 to 5 of 20 tasks) and Terminus~2 on Terminal-Bench~2.1 (+14.6\pp over its public baseline). Because Terminal-Bench~4 revises Terminal-Bench~3, the six benchmarks form five independent task sets; pooled over them, the improvement is significant ($p<0.001$), and it remains significant on the three sets released after the training data was collected ($p=0.004$). The trained model also takes fewer agent steps: the median falls from 150 to 98 on DeepSWE and from 102 to 78 on Terminal-Bench~3.

To see what changed, we paired base and trained trajectories on the same tasks (\S\ref{sec:analysis}). The base model's DeepSWE failures are mostly near-misses: the median failed run still passes 86\% of its target tests, and 84\% of failed runs keep every pass-to-pass test. In the pairs we read, all from tasks the trained model newly solves, the base model's failures recur in four modes that follow a standard engineering loop: lost requirements (understand), narrow testing (test), silent regressions (maintain), and weak ground truth (verify). On the same tasks, the trained model implements the entire specification, derives its tests from the requirement instead of from its own implementation, tracks what must keep working, and builds its own ground truth when none is provided. Each behavior is one the training tasks and reward favor. The evidence points to what transfers: not domain knowledge, which the base model largely had, but a way of finishing work.

\paragraph{Contributions.}
\begin{enumerate}[leftmargin=*,itemsep=1pt,topsep=2pt]
  \item Transfer from a small expert-built RL dataset: one epoch on 1,700 tasks improves pass@1 on all six external benchmarks we evaluated, under three harnesses, including four released after the data was collected and one whose multi-hour horizon has no counterpart in training (\S\ref{sec:results}).
  \item A paired trajectory analysis showing that the base model's DeepSWE failures are mostly last-mile near-misses and that, on tasks the trained model newly solves, the base model's failures recur in four modes that the trained model avoids (\S\ref{sec:analysis}).
  \item A partial-credit reward gated on regressions, with an account that links each behavior change to a property of the training tasks and reward (\S\S\ref{sec:data}--\ref{sec:method}, \S\ref{sec:analysis}).
\end{enumerate}

\section{Related Work}
\label{sec:related}

\paragraph{RL for software-engineering agents.}
RLVR was named and established on mathematics and instruction following~\citep{tulu3}; DeepSeek-R1 scaled rule-based rewards, optimized with GRPO~\citep{deepseekmath2024}, to mathematics and short-form code~\citep{deepseekr1}, and RL has since moved to long-horizon agentic software engineering. SWE-RL~\citep{swerl2025} learns from software-evolution data with a rule-based patch-similarity reward; SWE-Gym~\citep{swegym2025}, SWE-smith~\citep{swesmith2025}, and R2E-Gym~\citep{r2egym2025} build executable training environments from real repositories at scale; DeepSWE-Preview~\citep{deepsweagentica2025}, an agent unrelated to the DeepSWE benchmark, trains with RL alone in such environments; SkyRL~\citep{skyrl2025} provides an RL pipeline for long-horizon agents; and frontier open models such as Kimi K2~\citep{kimik2} combine large-scale agentic data synthesis with RL. These efforts mostly mine or synthesize tasks at scale and evaluate on SWE-bench-style benchmarks~\citep{swebench2024,swebenchverified2024} whose format matches the training environments. We train on a small set of expert-built tasks whose specifications enumerate individually gradable requirements, with partial credit gated on regressions, and measure transfer across benchmarks, harnesses, and horizons.

\paragraph{Benchmarks for agentic coding.}
SWE-bench~\citep{swebench2024} and its human-validated subset~\citep{swebenchverified2024} made issue resolution on real repositories the standard task; SWE-Bench~Pro~\citep{swebenchpro2025} raises difficulty with long-horizon, multi-file, human-verified problems. Terminal-Bench~\citep{terminalbench2026} evaluates agents on hard tasks in containerized command-line environments and, through the Harbor framework~\citep{harbor2026}, has become a continuously versioned benchmark (versions 2.0, 2.1, 3, and 4)~\citep{tbench21,tbench3,tbench4}. DeepSWE~\citep{deepswe2026} asks for original features in existing repositories, and SWE-Marathon~\citep{swemarathon2026} for complete systems over multi-hour horizons. Scores on all of these depend on the agent harness~\citep{terminalbench2026,swemarathon2026}, which fixes the tools, prompts, and control loop the model operates in. Common harnesses include SWE-agent and mini-swe-agent~\citep{sweagent2024,minisweagent2025}, OpenHands~\citep{openhands2025}, Terminus~2~\citep{terminalbench2026}, and commercial CLIs such as \mbox{Claude~Code~\citep{claudecode2025}}.

\paragraph{Partial-credit and rubric-based rewards.}
Rewards proportional to the fraction of unit tests passed are established in code RL~\citep{rltf2023}. Rubric-based rewards have been used to train instruction following~\citep{advancedif2025,mehta2026complex} and open-ended reasoning~\citep{rubricsasrewards2025}; \citet{mehta2026complex} reward the fraction of expert-written criteria satisfied. Proxy rewards can be gamed~\citep{skalse2022}, and partial credit opens a specific trade: satisfying new checks at the expense of existing behavior. Our reward keeps partial credit on the requested behavior and removes that trade with a gate that assigns zero reward whenever a pass-to-pass test fails.

\paragraph{Transfer from post-training.}
\citet{chu2025sft} find that RL generalizes to unseen rule and visual variants where supervised fine-tuning memorizes; \citet{huan2025math} find that RL on mathematics transfers to other domains better than supervised fine-tuning on the same problems; \citet{yue2025rl} argue that much of RLVR's pass@1 gain on reasoning tasks comes from concentrating probability on solutions the base model could already sample. Prior work from our group reports transfer across task distributions: post-training on long-horizon multi-tool office work, with no software-engineering tasks, produced gains on five external evaluations, including SWE-Bench~Pro from 20.5\% to 26.3\%~\citep{mehta2026crossbench,ritchie2026office}; single-turn constraint-following training produced its largest gains on multi-turn dimensions, led by +22.1\pp on MultiChallenge instruction retention~\citep{mehta2026complex,multichallenge2025}; and training on an enterprise simulation improved out-of-distribution pass rates~\citep{mehta2026corecraft}. Here training and evaluation are both coding, but the benchmarks, harnesses, and horizons differ.

\section{Training Data}
\label{sec:data}

The training set consists of 1,700 tasks from Surge AI's expert-built agentic coding collections, in the two formats most common in agentic coding evaluation. Every task pairs a specification the agent sees with hidden checks of the requested behavior.

\subsection{Two task families}
\label{sec:families}

\paragraph{Repository tasks (1,000).} Each task is a real repository at a pinned commit and a request for a specific, realistic change. Two groups of tests grade the result: hidden fail-to-pass tests encode the requested behavior, failing on the original commit and passing with the task author's reference patch; pass-to-pass tests guard existing behavior and pass on both, drawing on the repository's own suite and, in about half the tasks, on hidden tests written by the task author. The format is that of SWE-bench~\citep{swebench2024}, but each task is newly written by an expert, and its specification states concrete, individually gradable requirements: edge cases, exact interfaces, output formats, restrictions, and what must remain unchanged.

\paragraph{Terminal tasks (700).} Each task places the agent in a working containerized environment and asks for a concrete deliverable, in the style of Terminal-Bench~\citep{terminalbench2026}. Expert-written hidden verifiers inspect the final state of the environment and score a set of criteria.

\subsection{Properties that shape the training signal}
\label{sec:properties}

Four properties of the tasks determine what the reward can teach; \S\ref{sec:analysis} returns to each.
\begin{enumerate}[label=(P\arabic*),leftmargin=2.6em,itemsep=1pt,topsep=2pt]
  \item \textbf{Gradable requirements.} Each stated requirement maps to one or more checks, so the reward can credit partial progress (\S\ref{sec:reward}).
  \item \textbf{Hidden graders.} The agent sees the specification but never the fail-to-pass tests or verifiers that grade the requested behavior, so it cannot code against visible cases and has to decide for itself what to check.
  \item \textbf{Protected behavior.} Repository specifications state what must not change, and pass-to-pass tests enforce it; a single regression zeroes the reward.
  \item \textbf{No oracle.} The agent receives no reference solution, so any evidence that its result is correct has to be produced inside the rollout.
\end{enumerate}
Hidden graders and the absence of an oracle are common to executable coding environments. What is specific to these tasks is that their specifications enumerate individually gradable requirements, including what must not change, and that the reward gives partial credit per check, gated on regressions.

\subsection{Separation from the evaluation benchmarks}
\label{sec:separation}

All training tasks were collected before DeepSWE, Terminal-Bench~3 and~4, and SWE-Marathon were released. The two families mirror the formats of SWE-bench and Terminal-Bench, but neither was selected to improve a benchmark score. We checked every training task against all six benchmarks~\citep{swebenchpro2025,deepswe2026,tbench21,tbench3,tbench4,swemarathon2026}: no training task overlaps with or is derived from a benchmark task.

\section{Method}
\label{sec:method}

The recipe is about one RL epoch over the 1,700 tasks (108 rollout steps, 216 optimizer updates), with no supervised stage. The tasks follow a coarse two-stage curriculum, the 1,000 repository tasks first and the 700 terminal tasks after them, ordered by difficulty: the base model's pass rate was slightly higher on repository tasks. Prompts are shuffled within each stage.

\subsection{Reward}
\label{sec:reward}

Each rollout ends with the task's checks executed against the final environment state. Let $P_i\in\{0,1\}$ ($i\le N_P$) be the outcomes of the pass-to-pass tests (existing behavior) and $F_j\in[0,1]$ ($j\le N_F$) the outcomes of the fail-to-pass tests or verifier criteria (requested behavior), all weighted equally. The reward is
\begin{equation}
  R \;=\; \mathbf{1}\!\left[\,\sum_{i=1}^{N_P} P_i = N_P\right]\cdot \frac{1}{N_F}\sum_{j=1}^{N_F} F_j .
  \label{eq:reward}
\end{equation}
The second factor gives partial credit: a rollout that satisfies 3 of 12 requested checks scores 0.25. The first is a hard gate: a rollout that satisfies all 12 but fails 1 of 100 pass-to-pass tests scores 0. Terminal tasks have no pass-to-pass tests ($N_P=0$), so their reward is the fraction of verifier criteria satisfied. Rollouts truncated by a length limit are excluded from the loss, similar to DAPO's overlong filtering~\citep{dapo2025}.

\paragraph{Why partial credit matters under group-relative advantages.}
The reward enters the update only through advantages standardized within the $G$ rollouts of a prompt. Under a binary pass/fail reward, a group carries no signal whenever all $G$ rollouts fail or all pass. For a task on which each rollout passes with probability $q$, that happens with probability $q^G+(1-q)^G$, which for $G=8$ is 43\% at $q=0.1$ and 66\% at $q=0.05$: the harder the task, the more groups are silent. Under Equation~\eqref{eq:reward}, a group in which no rollout passes every check still ranks the rollouts that keep existing behavior by the fraction of target checks each satisfies. The gate places every regressing rollout at zero regardless of how many new requirements it meets, so there is no exchange rate between new and existing behavior for the policy to exploit. Both properties reappear as behaviors in \S\ref{sec:analysis}.

\subsection{Policy optimization}
\label{sec:policy}

We optimize the policy with GSPO~\citep{gspo2025}. For a prompt $x$ with rollouts $y_1,\dots,y_G$ and rewards $R_1,\dots,R_G$, GSPO uses the length-normalized sequence ratio and the group-standardized advantage,
\begin{equation}
  s_i(\theta)=\left(\frac{\pi_\theta(y_i\mid x)}{\pi_{\theta_\text{old}}(y_i\mid x)}\right)^{1/|y_i|},\qquad
  \hat A_i=\frac{R_i-\operatorname{mean}_j R_j}{\operatorname{std}_j R_j},
\end{equation}
with $\hat A_i=0$ when all rewards in the group are equal, and maximizes
\begin{equation}
  \mathcal{J}(\theta)=\mathbb{E}\left[\frac{1}{G}\sum_{i=1}^{G}\min\!\Big(s_i(\theta)\,\hat A_i,\ \operatorname{clip}\big(s_i(\theta),1-\varepsilon_{\text{low}},1+\varepsilon_{\text{high}}\big)\,\hat A_i\Big)\right].
\end{equation}
In a multi-turn rollout, $y_i$ comprises the tokens the model generates; tool outputs are masked out of the likelihoods and the loss. In mixture-of-experts models, the token-level ratios of GRPO~\citep{deepseekmath2024} fluctuate with expert routing, which motivates GSPO's sequence-level ratio~\citep{gspo2025}. The numerical mismatch between the rollout engine and the trainer is corrected with truncated importance sampling~\citep{tis2025}. There is no KL penalty and no entropy bonus. Each rollout step samples 16 prompts with $G=8$ rollouts each, and the 128 trajectories are consumed in two optimizer updates of 64. Groups that contain a rollout aborted by the infrastructure (for example, by a container crash) are discarded and replaced by groups sampled from additional prompts; groups whose rollouts all receive the same reward are kept and contribute no gradient.

\paragraph{Low-rank adaptation.} Instead of updating all parameters of the mixture-of-experts model, we train a rank-32 LoRA adapter~\citep{lora2022} with $\alpha=32$ on the attention and MLP projections, using Adam with decoupled weight decay~\citep{adam2015,adamw2019}. \citet{lorawithoutregret2025} find that LoRA applied to all weight matrices, the MLP layers included, matches full fine-tuning for RL even at small rank, because an RL episode carries little information.

\subsection{Rollouts and infrastructure}
\label{sec:infra}

Rollouts run in a minimal bash-only agent loop in the style of mini-swe-agent~\citep{minisweagent2025}, inside an isolated sandbox that provides the task's repository or environment and runs the task's checks when the rollout ends. The model runs in thinking mode at maximum reasoning effort with nucleus sampling (top-$p$ 0.95); each model turn is capped at 65,536 generated tokens, and a trajectory can use the model's full 262,144-token context window. Training runs on 64 NVIDIA H200 GPUs, with rollout generation and policy updates co-located. The stack is built from open-source components: Miles~\citep{miles2026}, an RL framework derived from slime~\citep{slime2025}, orchestrates SGLang~\citep{sglang2024} for generation and Megatron-LM with Megatron-Bridge~\citep{megatron2019,megatronbridge2025} and FlashAttention~\citep{flashattention2022} for training, combining tensor, pipeline, context, and expert parallelism to fit the 1T-parameter model. All results use the final checkpoint, fixed in advance; no hyperparameter was tuned on any benchmark.

\section{Evaluation Protocol}
\label{sec:eval}

\begin{table}[t]
  \caption{The six external benchmarks and how each departs from the training data. Training used repository and terminal tasks in a mini-swe-agent-style loop, with nothing at the horizon and scope of SWE-Marathon's builds. $^{*}$Released after the training data was collected.}
  \label{tab:benchmarks}
  \centering
  \footnotesize
  \setlength{\tabcolsep}{4pt}
  \begin{tabularx}{\linewidth}{@{}l r >{\hsize=1.3\hsize}Y l >{\hsize=0.7\hsize}Y@{}}
    \toprule
    Benchmark & Tasks & What it asks for & Harness & Departure from training \\
    \midrule
    SWE-Bench~Pro & 731 & Issue resolution in 11 repositories (public split); human-verified problems that require multi-file changes & mini-swe-agent & Same format \\
    \addlinespace[2pt]
    DeepSWE v1.1$^{*}$ & 113 & Original features in 91 repositories and five languages; reference solutions average 668 added lines & mini-swe-agent & Larger features \\
    \addlinespace[2pt]
    Terminal-Bench~2.1 & 89 & Hard command-line tasks in software engineering, system administration, scientific computing, security, and data science & Terminus~2 & Same format; unseen harness \\
    \addlinespace[2pt]
    Terminal-Bench~3$^{*}$ & 74 & Harder, longer tasks in seven domains (software, science, ML, operations, security, hardware, media); open internet & mini-swe-agent & Harder tasks; open internet \\
    \addlinespace[2pt]
    Terminal-Bench~4$^{*}$ & 66 & Terminal-Bench~3 with 8 tasks removed and 19 fixed; 8-hour agent limit & mini-swe-agent & As Terminal-Bench~3 \\
    \addlinespace[2pt]
    SWE-Marathon v1.1$^{*}$ & 20 & Complete systems (library reproductions, product clones, ML engineering, algorithmic optimization); agent limits up to 10 hours; expert estimates up to 400 hours & Claude~Code & Multi-hour horizon; unseen harness \\
    \bottomrule
  \end{tabularx}
\end{table}

\paragraph{Benchmarks.} Table~\ref{tab:benchmarks} lists the six benchmarks and their harnesses~\citep{swebenchpro2025,deepswe2026,tbench21,tbench3,tbench4,swemarathon2026}. Terminal-Bench~4 revises Terminal-Bench~3 (8 tasks removed, 19 fixed), so we treat the two as one family. These six are the only benchmarks we evaluated.

\paragraph{Harnesses.} Scores vary substantially across harnesses~\citep{terminalbench2026,swemarathon2026}, so we fix one per benchmark and use it unchanged for both models: mini-swe-agent~\citep{minisweagent2025} for SWE-Bench~Pro, DeepSWE (the leaderboard harness for all models), and Terminal-Bench~3 and~4 (the harness \citet{aa_tb4} uses for Terminal-Bench~4); Terminus~2~\citep{terminalbench2026} for Terminal-Bench~2.1, the harness of its public baseline; and, for SWE-Marathon, Claude~Code~\citep{claudecode2025}, one of the harnesses the benchmark authors evaluated, connected to the model through an API-compatible endpoint. Terminus~2 and Claude~Code differ from the training loop in tool interface, prompts, and control flow.

\paragraph{Metric and baselines.} We report pass@1 from a single run per benchmark with the final checkpoint, in thinking mode at maximum reasoning effort. In-house runs of both models use the same inference stack and sampling settings; only the LoRA adapter differs. Two baselines are publicly reported scores for the same checkpoint and harness: Terminal-Bench~2.1 (67.4\%) from \citet{aa_tb21}, which averages three runs under Terminus~2 (\citet{vals_tb21} independently reports 67.0\%), and DeepSWE from the benchmark leaderboard~\citep{deepswe2026}. Our in-house DeepSWE base run under the same harness scored 26.5\% (30 of 113), below the public 31.0\%, so the headline DeepSWE gain is conservative; the trajectory analysis in \S\ref{sec:analysis} uses this in-house run. All other baselines are in-house runs, which makes Terminal-Bench~2.1 the only benchmark without an in-house base run. Scoring rules and statistics are in Appendix~\ref{app:eval}.

\section{Results}
\label{sec:results}

\begin{table}[t]
  \centering
  \begin{threeparttable}
  \caption{Pass@1 (\%) of Kimi~K2.7~Code before and after RL on 1,700 expert-built coding tasks. Gains are differences of the rounded scores.}
  \label{tab:main}
  \small
  \begin{tabular}{@{}l l r r r r@{}}
    \toprule
    Benchmark & Harness & Tasks & Base & +RL & Gain (pp) \\
    \midrule
    SWE-Bench~Pro      & mini-swe-agent & 731 & 60.1 & 64.8 & +4.7 \\
    DeepSWE            & mini-swe-agent & 113 & 31.0\rlap{$^\dagger$} & 43.4 & +12.4 \\
    Terminal-Bench~2.1 & Terminus~2     &  89 & 67.4\rlap{$^\dagger$} & 82.0 & +14.6 \\
    Terminal-Bench~3   & mini-swe-agent &  74 & 1.4  & 12.1\rlap{$^\ddagger$} & +10.7 \\
    Terminal-Bench~4   & mini-swe-agent &  66 & 0.0  & 7.6  & +7.6 \\
    SWE-Marathon       & Claude~Code    &  20 & 5.0  & 25.0 & +20.0 \\
    \bottomrule
  \end{tabular}
  \begin{tablenotes}[flushleft]\footnotesize
    \item[$\dagger$] Publicly reported for the same checkpoint and harness (\S\ref{sec:eval}); our in-house DeepSWE base run scored 26.5.
    \item[$\ddagger$] Of the trained model's 74 runs, 8 failed for harness reasons and are not scored; 8 of the remaining 66 pass.
  \end{tablenotes}
  \end{threeparttable}
\end{table}

\subsection{The trained model improves on all six benchmarks}
\label{sec:main}

The gains in Table~\ref{tab:main} and Figure~\ref{fig:results} correspond to 35 more solved tasks on SWE-Bench~Pro, 14 on DeepSWE (19 against our in-house base run), 13 on Terminal-Bench~2.1, 7 on Terminal-Bench~3, 5 on Terminal-Bench~4, and 4 on SWE-Marathon. Without correction for multiple comparisons, the gains on Terminal-Bench~2.1, 3, and~4, and on DeepSWE against our in-house base run, are individually significant at the 0.05 level (Boschloo's exact test~\citep{boschloo1970}). A Cochran--Mantel--Haenszel test~\citep{mantelhaenszel1959} over the five independent task sets, with the Terminal-Bench~3/4 family counted once through Terminal-Bench~4, rejects the null of equal pass rates at $p<0.001$ (Appendix~\ref{app:stats}).

\subsection{Where the gains appear}
\label{sec:where}

The gains are not confined to the benchmarks closest to training (Table~\ref{tab:benchmarks}), SWE-Bench~Pro (+4.7\pp) and Terminal-Bench~2.1 (+14.6\pp). The four released after the training data was collected improve by 7.6 to 20.0\pp, and pooled over their three independent task sets the improvement is significant on its own ($p=0.004$). Both benchmarks evaluated under harnesses absent from training improve by more than 14\pp: Terminal-Bench~2.1 under Terminus~2, and SWE-Marathon, whose multi-hour builds have no counterpart in training, under Claude~Code. Only these two test transfer across harnesses: the other four run under mini-swe-agent, whose bash-only loop the training loop follows. The gains also span the difficulty range. On Terminal-Bench~2.1, where the base model passes two-thirds of the tasks, RL closes 45\% of the remaining gap; on Terminal-Bench~3 and~4, where the base model passes 1 of 74 and 0 of 66 tasks, the trained model passes 8 of 66 and 5 of 66.

\subsection{Fewer agent steps}
\label{sec:steps}

\begin{figure}[t]
  \centering
  \includegraphics[width=\linewidth]{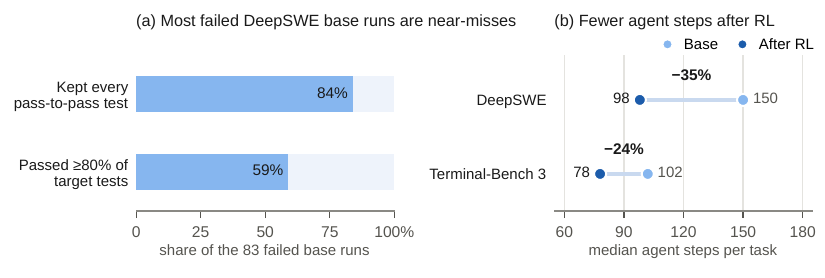}
  \caption{(a)~Failed base runs on DeepSWE are mostly near-misses: of the 83 failed in-house base runs, 84\% kept every pass-to-pass test and 59\% passed at least 80\% of their target tests; the median failed run passed 86\%. (b)~Median agent steps per task over all tasks, passed or failed; the DeepSWE base is the in-house run.}
  \label{fig:behavior}
\end{figure}

On the two benchmarks where we counted agent steps, the trained model reaches these scores in fewer steps (Figure~\ref{fig:behavior}b). Over all 113 DeepSWE tasks, the median trajectory falls from 150 agent steps for the in-house base run to 98 ($-$35\%), and over Terminal-Bench~3 from 102 to 78 ($-$24\%).

\section{What Changed: A Paired Trajectory Analysis}
\label{sec:analysis}

We paired base and trained trajectories on the same tasks and read the models' reasoning, shell commands, tests, and outputs, concentrating on tasks from DeepSWE, Terminal-Bench~3, and SWE-Marathon that the base model failed and the trained model passed. The cases below therefore come from tasks the trained model newly solves; the near-miss statistics in \S\ref{sec:nearmiss} cover every failed in-house DeepSWE base run.

\subsection{The base model's DeepSWE failures are mostly near-misses}
\label{sec:nearmiss}

Of the 83 failed in-house DeepSWE base runs, 59\% were near-misses that passed at least 80\% of their target tests (the tests of the requested behavior); the median failed run passed 86\%, and 84\% of failed runs preserved every pass-to-pass test (Figure~\ref{fig:behavior}a). In the pairs we read, the typical failure had most of the feature built and one or two requirements missing. A test fraction is not a requirement fraction: in the first case below, one dropped requirement fails 16 of 137 tests.

\subsection{Four failure modes and the behaviors after RL}
\label{sec:modes}

The base model's misses recur in four modes, and in each the trained model's trajectory on the same task shows a different behavior (Table~\ref{tab:modes}). The four follow a standard engineering loop (understand the requirement, test it, maintain what already works, verify against sound assumptions) and line up with the four task properties of \S\ref{sec:properties}.

\begin{table}[t]
  \caption{Failure modes of the base model in paired trajectories, the trained model's behavior on the same tasks, and the task property (\S\ref{sec:properties}) that favors the change.}
  \label{tab:modes}
  \centering
  \footnotesize
  \setlength{\tabcolsep}{4pt}
  \begin{tabularx}{\linewidth}{@{}l Y Y l@{}}
    \toprule
    Stage & Failure mode before RL & Behavior after RL & Property \\
    \midrule
    Understand & \textbf{Lost requirements.} Built most of the feature but dropped a required constraint, often a subtle one. & \textbf{Implements the entire specification.} Carries every stated requirement through implementation and final review. & (P1) Gradable requirements \\
    \addlinespace[2pt]
    Test & \textbf{Narrow testing.} Tested the cases its own incomplete implementation was built around. & \textbf{Tests the requirement, not the implementation.} Derives positive, negative, and alternate-form cases from the specification. & (P2) Hidden graders \\
    \addlinespace[2pt]
    Maintain & \textbf{Silent regressions.} Satisfied the new requirement while breaking behavior that had to stay intact. & \textbf{Protects existing behavior.} Tracks what must change and what must keep working before editing. & (P3) Protected behavior \\
    \addlinespace[2pt]
    Verify & \textbf{Weak ground truth.} Validated against a false assumption or a weak proxy. & \textbf{Reconstructs ground truth.} Builds an independent reference, an emulator, or its own test inputs. & (P4) No oracle \\
    \bottomrule
  \end{tabularx}
\end{table}

\subsubsection{Lost requirements \texorpdfstring{\ra}{\textrightarrow} implementing the entire specification}

\paragraph{Before.} A FastAPI task listed 20 requirements. Requirement 18 asked for a tracking middleware that exposes \code{get\_stats()} and \code{reset\_stats()} methods. The base model built nearly the entire feature, with the right architecture and working code, but defined the two functions at module level instead of as methods on the middleware object. It passed 121 of 137 target tests; the remaining 16 called the methods where the specification said they would be, and all 16 failed.

\paragraph{After.} The trained model made them instance methods and passed all 137 tests. The base model's understanding of the feature was not wrong: one interface requirement was lost between reading the specification and finishing the implementation, and the trained model carried it through.

\paragraph{Link to training (P1).} Every requirement maps to checks, so a rollout that drops one loses that requirement's share of the reward (on a task graded like this one, 16 of 137 checks). Partial credit keeps that pressure on when no rollout in a group finishes: of two rollouts that keep existing behavior, the one that passes more checks receives the higher advantage, where a pass/fail reward would give the whole group no signal (\S\ref{sec:reward}).

\subsubsection{Narrow testing \texorpdfstring{\ra}{\textrightarrow} testing the requirement, not the implementation}

\paragraph{Before.} Both models test often; they differ in what they choose to test. One task required default parameter values to work for both anonymous and named function definitions. The base model chose an implementation that handled the anonymous form, \code{func(a = 1)}, and every test it wrote used that form. The specification also required the named form, \code{func f(a = 1)}; the implementation did not handle an identifier between \code{func} and the parenthesis, and both hidden test suites failed. The base model had designed its tests around the path its implementation already supported.

\paragraph{After.} The trained model began with the named example from the specification itself, then tried anonymous functions, multi-line parameter lists, spread calls, immediately invoked functions, and other variants. It looked for cases that could prove its reading of the requirement wrong, not for confirmation that the cases it already supported worked.

\paragraph{Link to training (P2).} With the grader hidden, a rollout's own tests of the stated requirements are its best evidence about the checks that determine its reward. Rollouts that test what was asked, not only what was built, catch more of their own errors and score higher within their group, and the advantage credits the testing along with the fix.

\subsubsection{Silent regressions \texorpdfstring{\ra}{\textrightarrow} protecting existing behavior}

\paragraph{Before.} A task required eliminating layout shift across a site's pages while preserving the visible elements, styling, and analytics side effects. The base model reached zero layout shift on every page, partly by deleting two components. Its own final review judged the deletion harmless because some of the components' styling had been inlined elsewhere.

\paragraph{After.} Before editing, the trained model noted what the specification required to remain, then preserved those components while changing how their visual effects were applied. It worked from two lists: what had to change and what was not allowed to.

\paragraph{Link to training (P3).} On repository tasks, a rollout that fails any pass-to-pass test earns zero however many new requirements it meets, so it ties for last in its group and receives a negative advantage whenever another rollout earns credit.

\subsubsection{Weak ground truth \texorpdfstring{\ra}{\textrightarrow} reconstructing ground truth}

\paragraph{Before.} Some tasks offer no reference implementation or check to validate against. There the base model would sometimes adopt an assumption and then build a test that inherited it. On a two-dimensional spectral-solver task whose reference functions were unavailable inside the container, the base model decided the problem was effectively one-dimensional, built a one-dimensional solver, and validated it against reference data constructed under the same assumption. The test had no way to expose the assumption.

\paragraph{After.} The trained model used the method of manufactured solutions: it chose an analytic two-dimensional solution, derived the input for which that solution holds exactly, and used the pair as an independent check, which exposed an instability before submission. On a Verilog console task, it tested the hardware implementation differentially, against a Python emulator it wrote, instruction by instruction. On a SWE-Marathon task that required a Zstandard decompressor in an environment with no reference \code{zstd} binary, it first wrote a compressor, so that it could generate test inputs where none existed and check, round trip, that its decompressor recovered the original.

\paragraph{Link to training (P4).} The training set has no tasks on partial differential equation solvers, hardware emulators, or compressors, but every task puts the model in the same position: a precise specification, a hidden grader, and no oracle. Rollouts that produce their own evidence that a result is correct earn reward more consistently, which is the behavior these trajectories show. The SWE-Marathon authors report that frontier-agent failures often arise from poor self-verification~\citep{swemarathon2026}.

\subsection{A common thread}
\label{sec:thread}

In all four modes the base model stopped short of establishing that the whole job was done: it declared completion without checking every requirement, tested what it had built instead of what was asked, changed what it should have preserved, or verified against its own assumption. The trained model treats the specification as a contract to be met in full and its own work as a hypothesis to be tested. It adds this verification and still takes fewer agent steps on DeepSWE and Terminal-Bench~3 (\S\ref{sec:steps}). These are engineering habits, not the conventions of any one benchmark, which is consistent with gains that hold across benchmarks and harnesses.

\section{Conclusion}
\label{sec:conclusion}

One epoch of RL on 1,700 expert-built coding tasks, with GSPO and a rank-32 LoRA adapter, improved Kimi~K2.7~Code on all six external benchmarks we evaluated, by 4.7 to 20.0\pp, including on four benchmarks released after the data was collected, under two harnesses absent from training, and on multi-hour builds with no counterpart in training (1 to 5 of 20 tasks). On the newly solved tasks we read, the trained model implements the entire specification, tests the requirement, protects existing behavior, and builds its own ground truth, each a behavior the training tasks and reward favor. On this evidence, what transferred was not domain knowledge, which the base model largely had, but a way of finishing work.

\bibliographystyle{plainnat}
{\exhyphenpenalty=50 \raggedright 
\bibliography{references}}

\clearpage
\appendix

\section{Evaluation Details and Statistics}
\label{app:eval}

\subsection{Scoring}
\label{app:scoring}

All in-house results are single runs. On Terminal-Bench~3, runs that fail for harness reasons are not scored; for the trained model this excludes 8 of the 74 runs, and 8 of the remaining 66 pass. Counting those 8 runs as failures instead gives 10.8\% (8 of 74), a gain of 9.4\pp ($p=0.016$). The base model passes 1 task; any base denominator from 69 to 74 gives 1.4\% and $p\le0.013$ ($p\le0.021$ with the 8 runs counted as failures). On SWE-Marathon, one task requires an A100 GPU that was not available to us and is scored as failed for both models.

\subsection{Statistics}
\label{app:stats}

Table~\ref{tab:stats} gives pass counts, 95\% confidence intervals for each gain (hybrid score interval for a difference of independent proportions;~\citealp{newcombe1998}), and two-sided $p$-values from Boschloo's exact test~\citep{boschloo1970}. Public baselines enter as one run with the nearest count (35/113 for DeepSWE; 60/89 for Terminal-Bench~2.1, a mean of three runs), which is conservative for a multi-run mean. The pooled tests are Cochran--Mantel--Haenszel tests~\citep{mantelhaenszel1959} without continuity correction, stratified by benchmark; they use the public DeepSWE baseline and count the Terminal-Bench~3/4 family once, through Terminal-Bench~4, the more conservative choice (Terminal-Bench~3 instead gives $p=5\times10^{-4}$). Over the five independent task sets, $p=7\times10^{-4}$ (exact conditional version: $8\times10^{-4}$), and the Mantel--Haenszel risk difference is 7.0\pp (95\% CI [3.0, 11.0];~\citealp{greenland1985}). Over the three task sets released after the training data was collected (DeepSWE, Terminal-Bench~4, SWE-Marathon), $p=0.004$ (exact conditional version: 0.005). Substituting the in-house DeepSWE run gives $p=3\times10^{-4}$ over all five sets; dropping both benchmarks with public baselines gives $p=0.02$. The intervals and tests treat the two runs as independent samples, which is conservative when outcomes of the two checkpoints on the same task are positively correlated.

\begin{table}[h]
  \caption{Pass counts and uncertainty for Table~\ref{tab:main}: 95\% confidence interval of each gain and two-sided $p$-value of Boschloo's exact test.}
  \label{tab:stats}
  \centering
  \small
  \begin{tabular}{@{}l r r c r@{}}
    \toprule
    Benchmark & Base & +RL & 95\% CI of gain (pp) & $p$ \\
    \midrule
    SWE-Bench~Pro & 439/731 & 474/731 & [$-$0.2, 9.7] & 0.059 \\
    DeepSWE, public baseline & 35/113 & 49/113 & [$-$0.2, 24.4] & 0.055 \\
    DeepSWE, in-house baseline & 30/113 & 49/113 & [4.4, 28.5] & 0.008 \\
    Terminal-Bench~2.1 & 60/89 & 73/89 & [1.8, 26.8] & 0.028 \\
    Terminal-Bench~3 & 1/74 & 8/66 & [2.5, 20.8] & 0.010 \\
    Terminal-Bench~4 & 0/66 & 5/66 & [0.6, 16.5] & 0.029 \\
    SWE-Marathon & 1/20 & 5/20 & [$-$3.2, 42.3] & 0.089 \\
    \bottomrule
  \end{tabular}
\end{table}

\end{document}